\documentclass[10pt,twocolumn,letterpaper]{article}

\usepackage{cvpr}
\usepackage{times}
\usepackage{epsfig}
\usepackage{graphicx}
\usepackage{amsmath}
\usepackage{amssymb}
\usepackage{subfigure}
\usepackage{multirow}
\usepackage{authblk}
\usepackage{pdfpages}

\usepackage[breaklinks=true,bookmarks=false]{hyperref}

\cvprfinalcopy 

\def\cvprPaperID{6333} 

\begin{document}

\title{3D Human Mesh Regression with Dense Correspondence}

\makeatletter
\renewcommand\AB@affilsepx{ \protect\Affilfont}
\makeatother

\author[1]{Wang Zeng}
\author[2]{Wanli Ouyang}
\author[3]{Ping Luo}
\author[4]{Wentao Liu}
\author[1,4]{Xiaogang Wang}

\affil[1]{\small The Chinese University of Hong Kong} 
\affil[2]{The University of Sydney}
\affil[3]{The University of Hong Kong}
\affil[4]{SenseTime Research\authorcr}
\affil[ ]{\tt\small \{zengwang@link, xgwang@ee\}.cuhk.edu.hk, 
wanli.ouyang@sydney.edu.au, pluo@cs.hku.hk, liuwentao@sensetime.com}

\maketitle

\begin{abstract}
	Estimating 3D mesh of the human body from a single 2D image is an important task with many applications such as augmented reality and Human-Robot interaction.
	However, prior works reconstructed 3D mesh from global image feature extracted by using convolutional neural network (CNN), where the dense correspondences between the mesh surface and the image pixels are missing, leading to suboptimal solution.
	This paper proposes a model-free 3D human mesh estimation framework, named DecoMR, which explicitly establishes the dense correspondence between the mesh and the local image features in the UV space (\ie a 2D space used for texture mapping of 3D mesh).
	DecoMR first predicts pixel-to-surface dense correspondence map (\ie, IUV image), with which we transfer local features from the image space to the UV space. Then the transferred local image features are processed in the UV space to regress a location map, which is well aligned with transferred features.
	Finally we reconstruct 3D human mesh from the regressed location map with a predefined mapping function.
	We also observe that the existing discontinuous UV map are unfriendly to the learning of network. Therefore, we propose a novel UV map that maintains most of the neighboring relations on the original mesh surface.
	Experiments demonstrate that our proposed local feature alignment and continuous UV map outperforms existing 3D mesh based methods on multiple public benchmarks.
	Code will be made available at \url{https://github.com/zengwang430521/DecoMR}.
	
\end{abstract}

\section{Introduction}
\begin{figure}[t]
	\centering
	\includegraphics[width=0.9\linewidth]{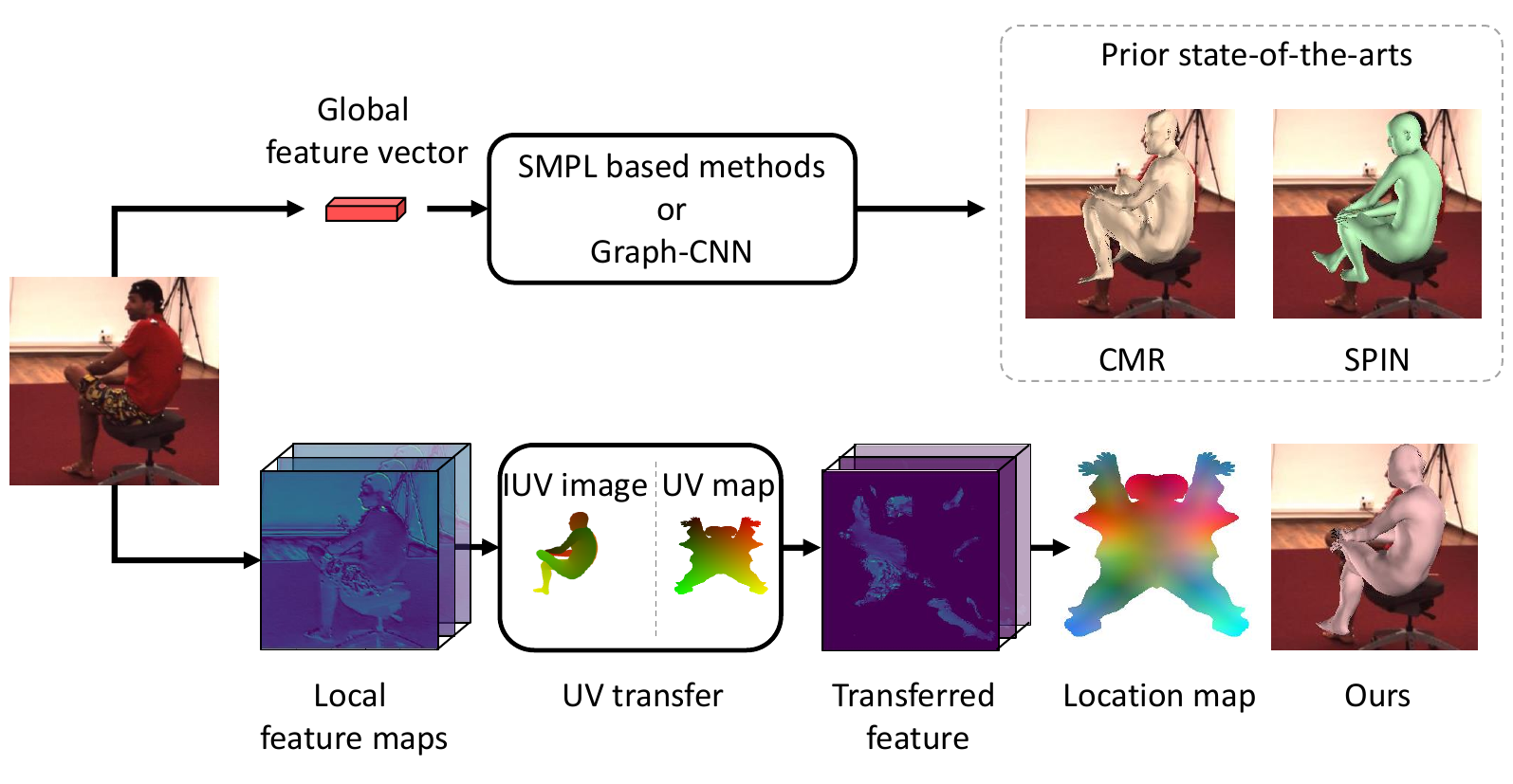}
	\caption{
		Prior methods (\eg, SPIN~\cite{kolotouros2019learning} and CMR~\cite{kolotouros2019convolutional}) usually reconstruct 3D meshes of human body from the global image feature vector extracted by neural networks, where the dense correspondences between the mesh surface and the image pixels are missing, leading to suboptimal results (top). Our DecoMR framework explicitly establishes such correspondence in the feature space with the aid of a novel continuous UV map, which results in better results in mesh details (bottom).
	}
	\label{fig:introduction}
	\vspace{-12pt}
\end{figure}

Estimation of the full human body pose and shape from a monocular image is a fundamental task for various applications such as human action recognition~\cite{hussein2013human, xia2012view}, VR/AR~\cite{huang2017towards} and video editing~\cite{huang2015hybrid}. 
It is challenging mostly due to the inherent depth ambiguity and the difficulty to obtain the ground-truth 3D human body data. 
There are several popular representations for 3D objects in literature, \eg, point clouds, 3D voxels and 3D meshes. Because of its compatibility with existing computer graphic engines and the efficiency to represent object surface in details with reasonable storage, 3D mesh representation has been widely adopted for 3D human body reconstruction~\cite{kanazawa2018end,bogo2016keep,kolotouros2019learning,guan2009estimating,pishchulin2016deepcut,zanfir2018monocular,huang2017towards,pavlakos2018learning,omran2018neural,yao2019densebody,kolotouros2019convolutional,zhu2019detailed}.


However, unlike 3D voxel representation, the dense correspondence between the template human mesh surface and the image pixels is missing, while this dense correspondence between the input and the output has been proven crucial for various tasks~\cite{newell2016stacked,zhu2019detailed}. 
Due to this limitation, most existing 3D mesh based methods, either model-based~\cite{kanazawa2018end, pavlakos2018learning, omran2018neural, kolotouros2019learning} or  model-free~\cite{kolotouros2019convolutional}, have to ignore the correspondence between the mesh representation and pixel representation. And they have to estimate the human meshes based on either global image feature~\cite{kanazawa2018end,kolotouros2019convolutional,kolotouros2019learning}, or hierarchical projection and refinement~\cite{zhu2019detailed}, which is time consuming and sensitive to initial estimation. 


To utilize the 3D mesh representation without losing the correspondence between the mesh space and the image space,
we propose a 3D human mesh estimation framework that explicitly establishes the dense correspondence between the output 3D mesh and the input image in the UV space. 

\emph{Representing output mesh by a new UV map:}
Every point on the mesh surface is represented by its coordinates on the continuous UV map. 
Therefore, the 3D mesh can be presented as a location map in the UV space, of which the pixel values are the 3D coordinates of its corresponding point on the mesh surface, as shown in Figure~\ref{fig:introduction}.
Instead of using SMPL default UV map, we construct a new continuous UV map that maintains more neighboring relations of the original mesh surface, by parameterizing the whole mesh surface into a single part on the UV plane, as shown in Figure~\ref{fig:introduction}.



\emph{Mapping image features to the UV space:}
To map the image features to the continuous UV map space, we first use a network that takes a monocular image as input for predicting an IUV image~\cite{alp2018densepose}, which assign each pixel to a specific body part location. 
Then the local image features from the decoder are transferred to the UV space with the guidance of predicted IUV image to construct the transferred feature maps that are well aligned with the corresponding mesh area.  

Given the transferred local features, we use both the local features and the global feature to estimate the location map in the UV space, which is further used to reconstruct the 3D human body mesh with the predefined UV mapping function.
Since our UV map is continuous and maintains the neighboring relationships among body parts, details between body parts can be well preserved when the local features are transferred.

In summary, our contributions are twofold:
\begin{enumerate}
	\item[$\bullet$] We propose a novel UV map that maintains most of the neighboring relations on the original mesh surface.
	
	\item[$\bullet$] We explicitly establish the dense correspondence between the output 3D mesh and the input image by the transferred local image features.
\end{enumerate}
We extensively evaluate our methods on multiple widely used benchmarks for 3D human body reconstruction. Our method achieves state-of-the-art performance on both 3D human body mesh reconstruction and 3D human body pose estimation. 


\section{Related Work}

\subsection{Optimization-based methods}
Pioneer works solve the 3D human body reconstruction by optimizing parameters of an predefined 3D human mesh models, \eg, SCAPE~\cite{anguelov2005scape} and SMPL~\cite{loper2015smpl}, with respect to the ground-truth body landmark locations~\cite{guan2009estimating}, or employing a 2D keypoints estimation network ~\cite{bogo2016keep}. 
To improve the precision, extra landmarks are used in~\cite{lassner2017unite}.
Recent work~\cite{zanfir2018monocular} enables multiple persons body reconstruction by incorporating human semantic part segmentation clues, scene and temporal constrains.

\subsection{Learning-based methods}
\textbf{Model-based methods:} 
Directly reconstruction of the 3D human body from a single image is a relatively hard problem. Therefore, many methods incorporate a parameterized 3D human model and change the problem into the model parameter regression. 
For example, HMR~\cite{kanazawa2018end} regresses the SMPL parameters directly from RGB image. 
In order to mitigate the lack of robustness caused by the inadequacy of in-the-wild training data, some approaches employ intermediate representations, such as 2D joint heatmaps and silhouette~\cite{pavlakos2018learning}, semantic segmentation map~\cite{omran2018neural} or IUV image~\cite{xu2019denserac}.
Recently, SPIN~\cite{kolotouros2019learning} incorporates 3D human model parameter optimization into network training process by supervising network with optimization result, and achieves the state-of-the-art results among model-based 3D human body estimation approaches.

Compared with optimization-based methods, model parameter regression methods are more computationally efficient. While these methods can make use of the prior knowledge embedded in 3D human model, and tend to reconstruct more biologically plausible human bodies compared with model-free methods, the representation capability is also limited by the parameter space with these predefined human models. 
In addition, as stated in \cite{kolotouros2019convolutional}, 3D human model parameter space might not be so friendly to the learning of network.
On the contrary, our framework does not regress model parameters. Instead, it directly outputs 3D coordinates of each mesh vertex.

\begin{figure*}[htp]
	\centering
	\includegraphics[width=1\linewidth]{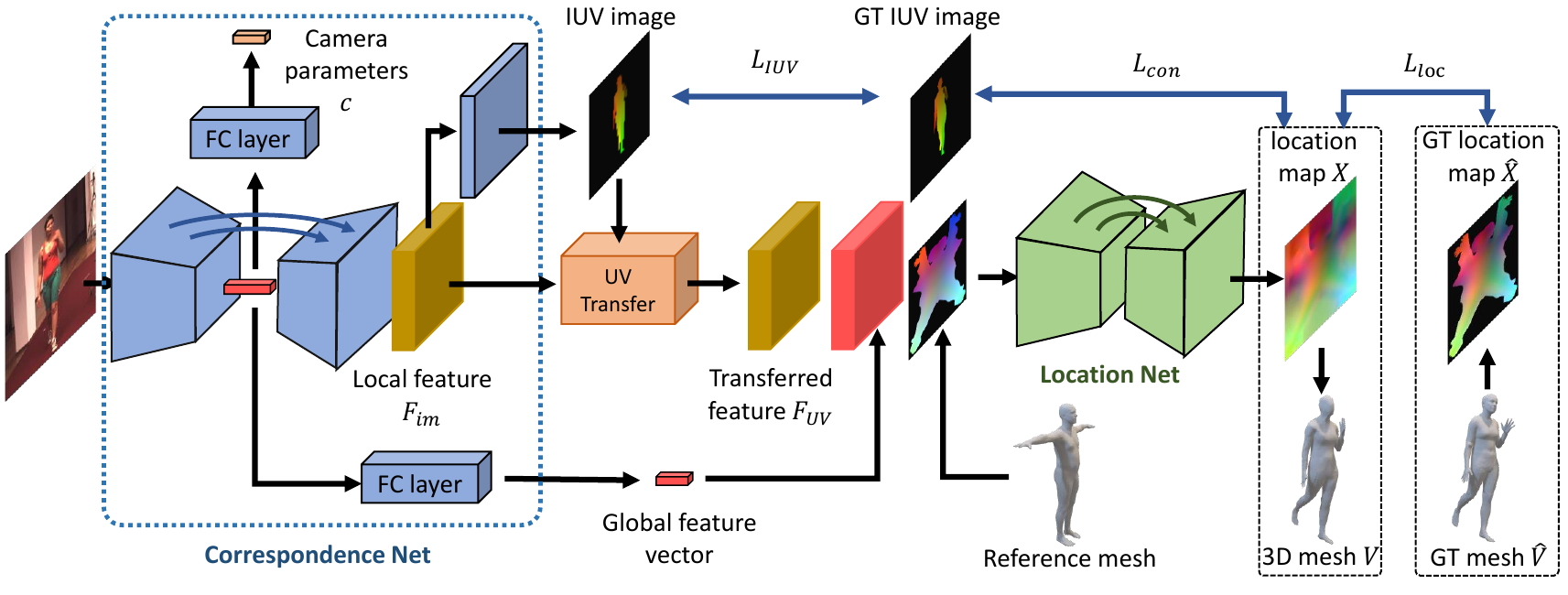}
	\caption{
		Overview of our framework. Given an input image, an IUV map is first predicted by the correspondence net. Then local image features are transferred to the UV space. Location net takes transferred local features, expanded global feature and reference location map as input, and regresses a location map. Finally, 3D mesh is reconstructed from the location map.
	}
	\label{framework}
	\vspace{-10pt}
\end{figure*}

\textbf{Model-free methods: }
Some methods do not rely on human models and regress 3D human body representation directly from image.
BodyNet~\cite{varol2018bodynet} estimates volumetric representation of 3D human with a Voxel-CNN. A recent work~\cite{gabeur2019moulding} estimates visible and hidden depth maps, and combines them to form a point cloud of human. 
Voxel and point cloud based representations are flexible and can represent objects with different topology. However, the capability of reconstructing surface details is limited by the storage cost. 

CMR~\cite{kolotouros2019convolutional} uses a Graph-CNN to directly regress 3D coordinates of vertices from image features. Densebody~\cite{yao2019densebody} estimates vertex location in the form of UV-position map.
A recent work~\cite{pumarola20193dpeople} represents the 3D shapes using 2D geometry images, which can be regarded as a special kind of UV-position map.
These methods do not use any human model.
However, they still lack correspondence between human mesh and image and estimate the whole surface only relying on global image feature. On the contrary, our method can employ local feature for the reconstruction of corresponding surface area.

The efficacy of the UV space representation has been demonstrated in recent work Tex2Shape~\cite{alldieck2019tex2shape}, where the 3D human shape is estimated from the texture map which is obtained by transferring images pixels according to the IUV image estimated by DensePose~\cite{alp2018densepose}. 
We also use the IUV image to guide the human mesh estimation. However, in ~\cite{alldieck2019tex2shape}, the UV transfer is used to preprocess the raw image and is independent from the model learning, while we incorporate the UV transfer into our network to enable the end-to-end learning.
We observe the efficacy of learning the transferred features end-to-end, which has also been proved by prior works, \eg, Spatial Transformer Networks~\cite{jaderberg2015spatial} and Deformable ConvNets~\cite{dai2017deformable}.

Very recently, HMD~\cite{zhu2019detailed} refines initial estimated human mesh by hierarchical projection and mesh deformation.
PIFu~\cite{saito2019pifu} reconstructs 3D human as implicit function.
HMD and PIFu are able to utilize local image features to achieve impressive details in the reconstruction results.
However, HMD is computationally intensive and sensitive to the initial estimation, while implicit function lacks the semantic information of human body.
In contrast, we estimate the pixel-to-surface dense correspondence from images directly, which is computationally efficient and more robust, and the location map maintains the semantic information of human body.
%


\section{Our Method}



\textbf{Overview.} As shown in Figure \ref{framework}, our framework DecoMR consists of two components, including a dense correspondence estimation network (CNet), which preforms in the image space, as well as a localization network (LNet), which performs on a new continuous UV map space.
The CNet has an encoder-decoder architecture to estimate an IUV image. It also extracts local image features $\mathcal{F}_{im}$, and then uses the the estimated IUV image for transferring the image features $\mathcal{F}_{im}$ to the transferred local features $\mathcal{F}_{UV}$ in the UV space. 
LNet takes the above transferred local features $\mathcal{F}_{UV}$ as input, and regresses a location map $X$, whose pixel value is the 3D coordinates of the corresponding points on the mesh surface.
Finally, the 3D human mesh $V$ is reconstructed from the above location map by using a predefined UV mapping function.
As a result, the location map and the transferred feature map are well aligned in the UV space, 
thus leading to dense correspondence between the output 3D mesh and the input image. 

Although the SMPL UV map~\cite{loper2015smpl} is widely used in the literature~\cite{yao2019densebody,alldieck2019tex2shape,grigorev2019coordinate}, it loses the neighboring relationships between different body parts as shown in Figure~\ref{fig:uv_map} (a),{ which is crucial for network learning as stated in \cite{kolotouros2019convolutional}.} 
Therefore, we design a new UV map that is able to maintain more neighboring relationships on the original mesh surface as shown in Figure~\ref{fig:uv_map} (b).

%

The overall objective function of DecoMR is  
\begin{equation}
\mathcal{L} = \mathcal{L}_{IUV} + \mathcal{L}_{Loc} + \lambda_{con}\mathcal{L}_{con}.
\label{eq:L}
\end{equation}
It has three loss functions of different purposes.
The first loss denoted as $\mathcal{L}_{IUV}$ minimizes the distance between the predicted IUV image and the ground-truth IUV image.
The second loss function denoted as $\mathcal{L}_{Loc}$ minimizes the dissimilarity between the regressed human mesh (\eg location map) and the ground-truth human mesh. 
%
In order to encourage the output mesh to be aligned with the input image, we add an extra loss function, denoted as $\mathcal{L}_{con}$, which is a consistent loss to increase the consistency between the regressed location map and the ground-truth IUV image.
The $\lambda_{con}$ in Equation \ref{eq:L} is a constant coefficient to balance the consistent loss $\mathcal{L}_{con}$.
We first define the new UV map below and then introduce different loss functions in details. 
%
%

\subsection{The Continuous UV map} \label{section_uv}

\begin{figure}[t]
	\centering
	\includegraphics[width=0.9\linewidth]{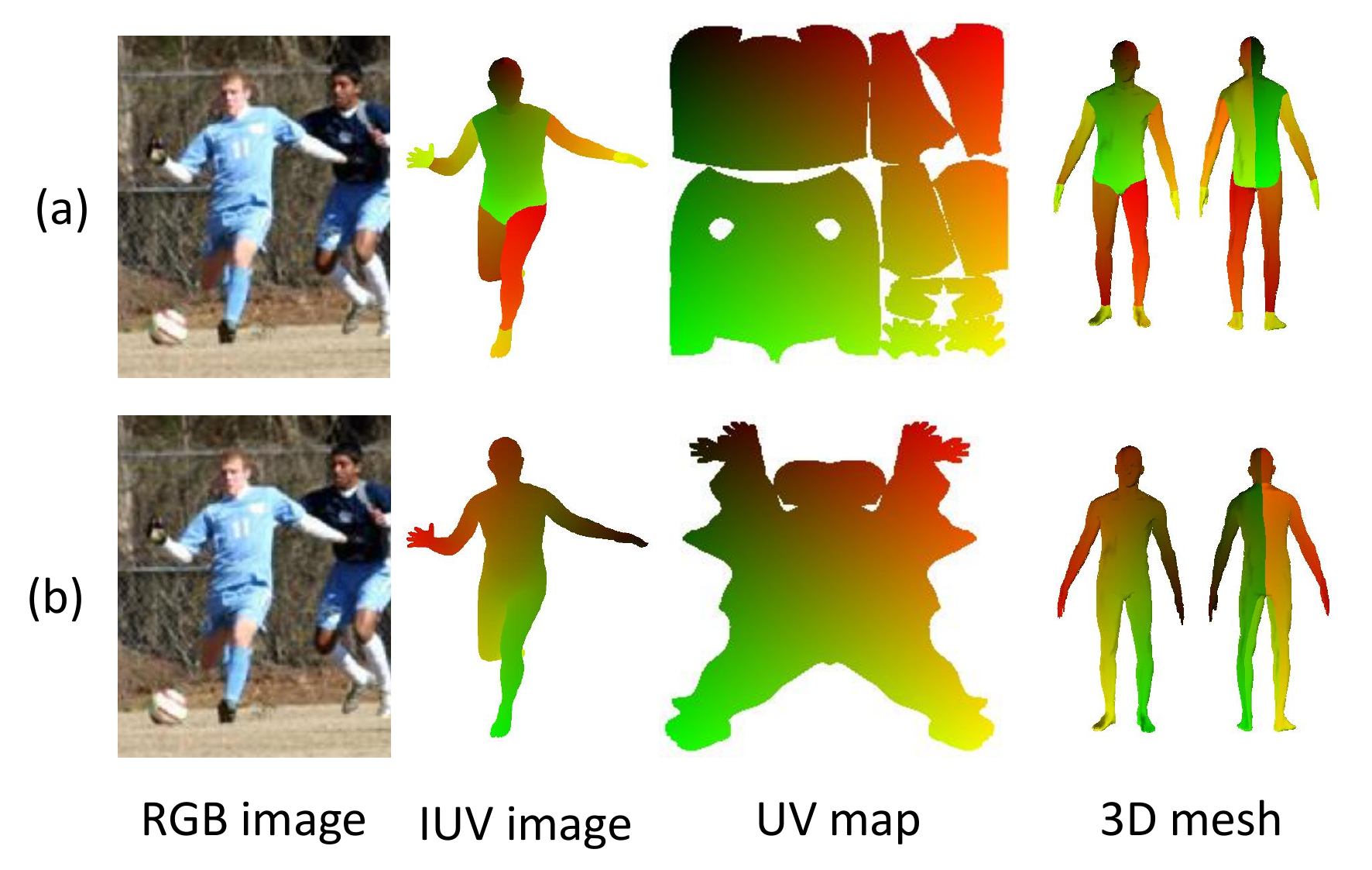}
	\caption{Comparisons of UV maps. Row (a) shows SMPL default UV map and row (b) shows our continuous UV map.}
	\label{fig:uv_map}
	\vspace{-5pt}
\end{figure}

%
First we define a new continuous UV map that preserves more neighboring relationships of the original mesh than the ordinary UV map of SMPL.
As shown in Figure~\ref{fig:uv_map} (a), multiple mesh surface parts are placed separately on the SMPL default UV map, which loses the neighboring relationships of the original mesh surface. 
Instead of utilizing SMPL UV map as \cite{alldieck2019tex2shape,grigorev2019coordinate,yao2019densebody}, 
we design a new continuous UV map. We first carefully split the template mesh into an open mesh, 
while keeping the entire mesh surface as a whole. 
Then we utilize an algorithm of area-preserving 3D mesh planar parameterization ~\cite{jacobson2017libigl,jiang2017simplicial}, to minimize the area distortion between the UV map and the original mesh surface, in order to obtain an initial UV map.
To maintain symmetry for every pair of symmetric vertices on the UV map, we further refine the initial UV map by 
first aligning the fitted symmetric axis with $v$ axis and then averaging the UV coordinates with the symmetric vertex flipped by $v$ axis.

\begin{figure}[t]
	\centering
	\includegraphics[width=0.9\linewidth]{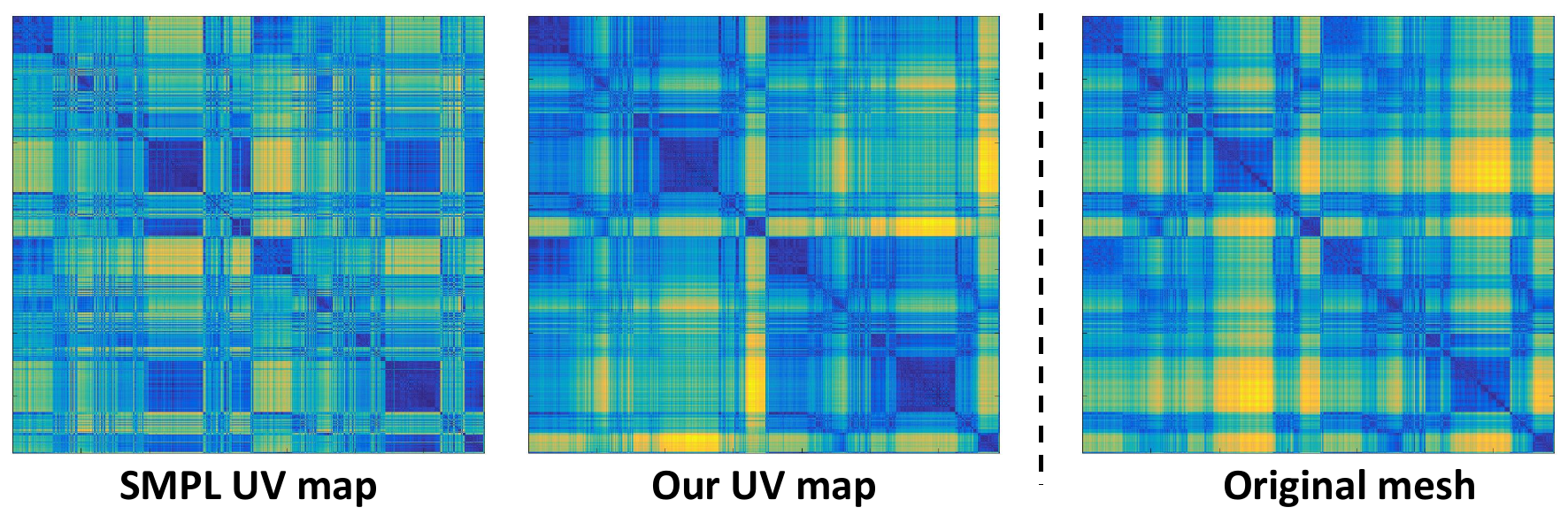}
	\caption{Comparisons of distance matrices between vertices calculated on SMPL UV map ,
		the proposed UV map, and the original mesh surface. Compared to SMPL UV map, the distance matrix of the proposed UV map is more similar to that of the original mesh. 
	}
	\label{distance}
\end{figure}

\textbf{Comparisons.} 
Here we quantitatively show that our continuous UV map outperforms the SMPL UV map in terms of preserving connection relationships between vertices on the mesh.
%
To do so, we compute the distance matrix, where each element is the distance between every vertex pair. We also compute the distance matrix on the UV map.
Figure~\ref{distance} shows such distance matrices.
This distance matrix can be computed by using different types of data. 
%
For the mesh surface, the distance between two vertices is defined as the length of the minimal path between them on the graph built from the mesh.
For the UV map, the distance between two vertices is directly calculated by the the distance between their UV coordinates.

Now we quantitatively evaluate the similarity between the distance matrices of UV map and original mesh in two aspects as shown in Table~\ref{tab:similarity}.
In the first aspect, we calculate the 2D correlation coefficient denoted as $S_1$. We have
\begin{equation}
S_1 = \frac{\sum\limits_{m} \sum\limits_{n}\left(A_{m n}-\bar{A}\right)\left(B_{m n}-\bar{B}\right)}{\sqrt{\left(\sum\limits_{m} \sum\limits_{n}\left(A_{m n}-\bar{A}\right)^{2}\right)\left(\sum\limits_{m} \sum\limits_{n}\left(B_{m n}-\bar{B}\right)^{2}\right)}},
\end{equation}
where $A$ and $B$ are the distance matrices of original mesh and UV map, respectively. $\bar{A}$ and $\bar{B}$ are the mean value of $A$ and $B$ respectively. $m$ and $n$ are the indices of mesh vertices.

In the second aspect,  
we calculate the normalized cosine similarity between the distance matrices of UV map and original mesh, denoted as $S_2$.
From Table~\ref{tab:similarity}, we see that our continuous UV map outperforms SMPL UV map by large margins on both metric values, showing that our UV map preserves more neighboring relationships than the SMPL UV map.

\begin{table}[t]
	\centering
	\small
	\begin{tabular}{c|c|c}
		\hline
		UV map &  2D correlation ($S_1$) & cosine similarity ($S_2$)\\
		\hline
		SMPL~\cite{loper2015smpl} &  0.2132 & 0.8306 \\
		Ours & 0.7758 & 0.9458 \\
		\hline
	\end{tabular}
	\caption{
		Comparisons of the similarity between the vertices' distance matrices of the original mesh surface and different types of UV maps.		
		$S_1$ is the 2D correlation coefficient and $S_2$ is the normalized cosine similarity. We see that the proposed UV map outperforms SMPL default UV map on both metrics.}
	\label{tab:similarity}
	\vspace{-10pt}
\end{table}

\textbf{Pixel-to-Mesh Correspondence.} With the proposed UV map, every point on the mesh surface can be expressed by its coordinates on the UV map (\ie UV coordinates).
Therefore, we can predict the pixel-to-surface correspondence by estimating the UV coordinates for each pixel belonging to human body, leading to an IUV image as shown in Figure~\ref{fig:uv_map}.
More importantly, we can also represent a 3D mesh with a location map in the UV space, where the pixel values are 3D coordinates of the corresponding points on the mesh surface. Thus it is easy to reconstruct 3D mesh from a location map with the following formula,
\begin{equation}
V_{i} = X(u_{i}, v_{i}),
\end{equation}
where $V_{i}$ denotes 3D coordinates of vertex, $X$ is the location map, $u_{i}$ and $v_{i}$ are UV coordinates of the vertex.

\subsection{Dense Correspondence Network (CNet)}\label{section_corr}

CNet establishes the dense correspondence between pixels of the input image and areas of 3D mesh surface.
As illustrated in Figure~\ref{framework}, CNet has an encoder-decoder architecture, where the encoder employs ResNet50~\cite{he2016deep} as backbone, and the decoder consists of several upsampling and convolutional layers with skip connection with encoder.
In particular, the encoder encodes the image as a local feature map and a global feature vector, as well as regresses the camera parameters, which are used to project the 3D mesh into the image plane.
The decoder first generates a mask of the human body, which distinguishes fore pixels (\ie human body) from those at the back. 
%
Then, the decoder outputs the exact UV coordinates for the fore pixels, constituting an IUV image as shown in Figure~\ref{fig:uv_map}. 
With the predicted IUV image, the corresponding point on the mesh surface for every image pixel can be determined. 
The loss function for the CNet contains two terms,
\begin{equation}
\mathcal{L}_{IUV}=\lambda_{c}\mathcal{L}_{c} + \lambda_{r}\mathcal{L}_{r},
\end{equation}
where $\mathcal{L}_{c}$ is a dense binary cross-entropy loss for classifying each pixel as `fore' or `back', $\mathcal{L}_{r}$ is an $l_1$ dense regression loss for predicting the exact UV coordinates, and $\lambda_{c}$ and $\lambda_{r}$ are two constant coefficients.

\begin{figure}[t]
	\centering
	\includegraphics[width=0.9\linewidth]{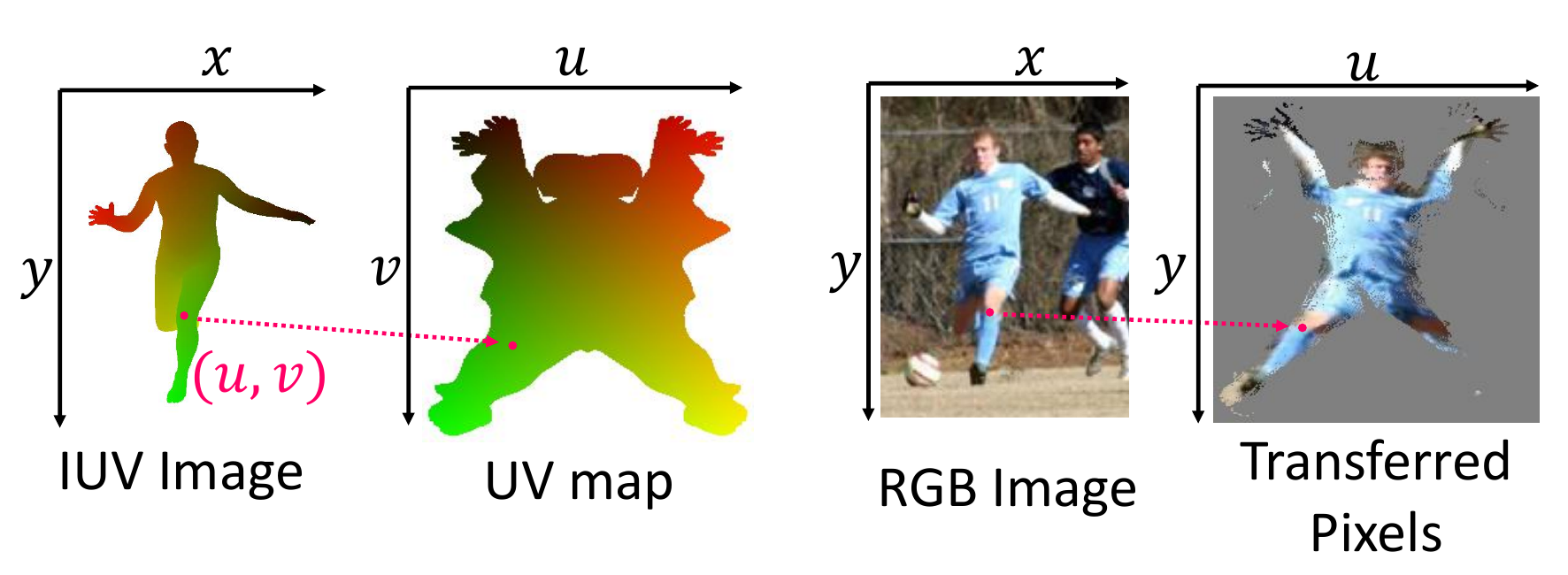}
	\caption{Illustration of the UV transferring of raw image pixels. Elements in the image space can be transferred to the UV space with the guidance of IUV image. }
	\label{uv_transfer}
	\vspace{-10pt}
\end{figure}

\subsection{Vertex coordinates regression} \label{section_loc}
The location net (LNet) aims to regress 3D coordinates of mesh vertices by outputting a location map, from which the 3D mesh can be reconstructed easily.
As shown in Figure~\ref{framework}, the LNet first transfers image features from the image space to the UV space with the guidance of predicted IUV image:
\begin{equation}
\mathcal{F}_{UV}(u, v) = \mathcal{F}_{im}(x, y), 
\label{equation:UV_transfer}
\end{equation}
where $(x, y)$ are the coordinates in image space of the pixels classified as fore, and $(u, v)$ are the predicted coordinates in UV space of these pixels. $\mathcal{F}_{im}$ is the feature map in image space and $\mathcal{F}_{UV}$ is the transferred feature map in UV space.

The feature map $\mathcal{F}_{UV}$ is well aligned with the output location map. So the LNet can predict location map utilizing corresponding local image features. In this way, the dense correspondence between image pixels and mesh surface areas is established explicitly.
An example of raw image pixels transferred to UV space is shown in Figure~\ref{uv_transfer}. Note that our framework transfers features instead of pixel values.

The LNet is a light CNN with skip connections taking the transferred local image features, expanded global image feature and a reference location map as input. 
Intuitively, we apply an weighted $l_1$ loss between the predicted location map $X$ and ground-truth location map $\hat{X}$, \ie,
\begin{equation}
\mathcal{L}_{map}=\sum_{u}\sum_{v}W(u,v)\cdot\left\|X(u,v)-\hat{X}(u,v)\right\|_{1}.
\vspace{-5pt}
\end{equation}
$W$ is a weight map used to balance the contribution of different mesh areas, where areas away from torso are assigned higher weights.

We also reconstruct a 3D human mesh from the predicted location map and 
get 3D joints from human mesh employing joint regressor as previous works~\cite{kanazawa2018end,kolotouros2019convolutional,kolotouros2019learning}. Then we add supervision on the 3D coordinates and projected 2D coordinates in the image space of the joints, \ie,
\vspace{-5pt}
\begin{equation}
\mathcal{L}^{3D}_{J}=\sum_{i}^{k}\left\|Z_{i}-\hat{Z}_{i}\right\|_{1},
\vspace{-5pt}
\end{equation}
\begin{equation}
\mathcal{L}_{J}^{2D}=\sum_{i}^{k}\left\|v_{i}(z_{i}-\hat{z}_{i})\right\|_{2}^{2},
\vspace{-5pt}
\end{equation}
where $Z_{i}$ and $z_{i}$ are the regressed 3D and 2D coordinates of joints, while $\hat{Z}_{i}$ and $\hat{z}_{i}$ refer to the coordinates of the ground-truth joints, and $v_i$ denotes the visibility of joints. 


Finally, the full loss for LNet is 
\begin{equation}
\mathcal{L}_{loc}= \mathcal{L}_{map} + \mathcal{L}^{3D}_{J} + \mathcal{L}_{J}^{2D}.
\vspace{-5pt}
\end{equation}

\begin{figure}[t]
	\centering
	\includegraphics[width=0.9\linewidth]{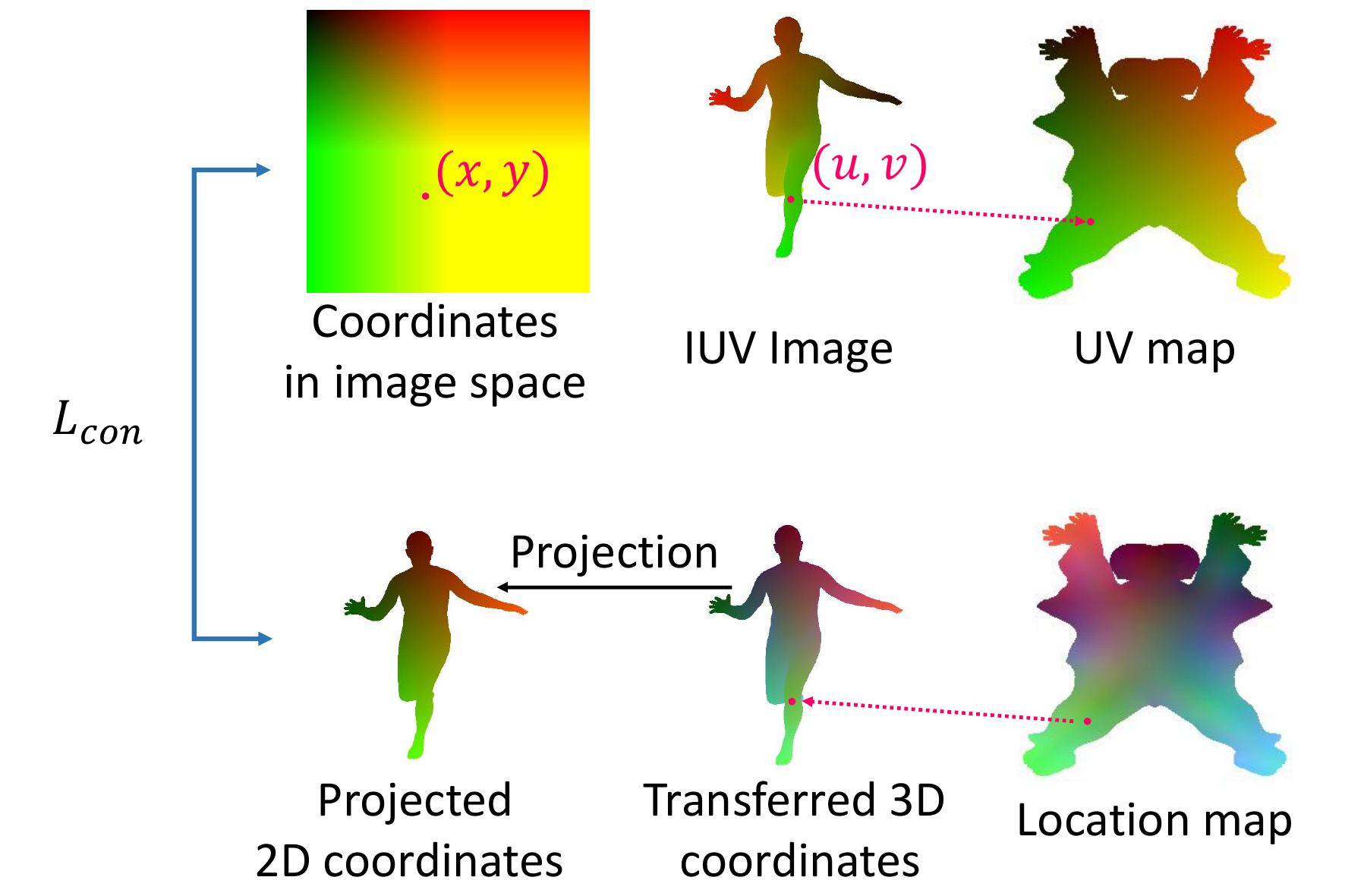}
	\caption{Illustration of our consistent loss between the location map and the IUV image. 
		3D coordinates in the location map are transferred back to the image space using IUV image, and then projected to the image plane. The projected 2D coordinates are supervised by the coordinates of image pixels in the image space.}
	\label{consistent}
	\vspace{-10pt}
\end{figure}

\textbf{Consistent Loss}:
Besides the above widely used supervision, we add an extra supervision between regressed location map and ground-truth IUV image to improve the alignment between 3D mesh and image.

As shown in Figure~\ref{consistent}, with an IUV image, we can also transfer location map from the UV space back to the image space and get 3D coordinates for every foreground pixel. The 3D coordinates are then projected to image plane to get 2D coordinates, which should be consistent with the coordinates of the pixels in the image space.
Then the consistent loss is constructed as follows:
\begin{equation}
\mathcal{L}_{con}=\sum_{(x, y)}
\left\|(x, y)-\pi(X(u, v), c))\right\|^{2}_{2},
\vspace{-5pt}
\end{equation}
where $X$ is the predicted location map, $\pi(X, c)$ denotes the projection function with predicted camera parameters $c$, and $x, y, u, v$ are the same as that in Equation \ref{equation:UV_transfer}.
This consistent loss is similar to the loss item $\mathcal{L}_{dense}$ in recent work of Rong \etal~\cite{Rong_2019_ICCV}. However, in our framework there is no need to calculate the corresponding point on mesh surface as in \cite{Rong_2019_ICCV}, because the correspondence between mesh surface and image pixel is already established.

\subsection{Implementation details} \label{section_detail}
We set $\lambda_{c}$, $\lambda_{r}$ and $\lambda_{cons}$ to $0.2$, $1$ and $1$ respectively and optimize the framework with an Adam optimizer~\cite{kingma2014adam}, with batch size 128 and learning rate 2.5e-4. The training data is augmented with randomly scaling, rotation, flipping and RGB channel noise. We first train the CNet for 5 epochs and then train the full framework end-to-end for 30 epochs.

\section{Experiments}

\subsection{Datasets}\label{dataset}

In the experiment, we train our model on the Human3.6M~\cite{ionescu2013human3}, UP-3D~\cite{lassner2017unite} and SURREAL~\cite{varol17_surreal} dataset, while we provide evaluations on the test set of Human3.6M, SURREAL and LSP dataset~\cite{Johnson10}.

\textbf{Human3.6M}: Human3.6M~\cite{ionescu2013human3} is a large scale indoor dataset for 3D human pose estimation, including multiple subjects performing typical actions like walking, sitting and eating.
Following the common setting~\cite{kanazawa2018end}, we use subjects S1,  S5, S6, S7 and S8 as training data and use subjects S9 and S11 for evaluation. 
For evaluation, results are reported using two widely used metrics (MPJPE and MPJPE-PA) under two popular protocols: P1 and P2, as defined in \cite{kanazawa2018end},




\textbf{UP-3D}: UP-3D~\cite{lassner2017unite} is an outdoor 3D human pose estimation dataset. It provides 3D human body ground truth by fitting SMPL model on images from 2D human pose benchmarks. 
We utilize the images of training and validation set for training.

\textbf{SURREAL}: SURREAL dataset~\cite{varol17_surreal} is a large dataset providing synthetic images with ground-truth SMPL model parameters. We use the standard split setting~\cite{varol17_surreal} but remove all images with incomplete human body and evaluate on the same sampled test set as BodyNet~\cite{varol2018bodynet}.

\textbf{LSP}: LSP~\cite{Johnson10} dataset is a 2D human pose estimation benchmark. In our work, we evaluate the segmentation accuracy of each model on the segmentation annotation~\cite{lassner2017unite}.

\begin{table}
	\begin{center}
		\begin{tabular}{c|c}
			\hline
			Methods & MPJPE-PA \\
			\hline
			Lassner \etc~\cite{lassner2017unite} & 93.9 \\
			SMPLify~\cite{bogo2016keep} &  82.3 \\
			\hline
			Pavlakos \etc~\cite{pavlakos2018learning} & 75.9 \\
			HMR\cite{kanazawa2018end} & 56.8 \\
			NBF\cite{omran2018neural} & 59.9 \\
			CMR\cite{kolotouros2019convolutional} & 50.1 \\
			DenseRaC\cite{xu2019denserac} & 48.0 \\
			SPIN\cite{kolotouros2019learning} & 41.1 \\ 
			\hline
			Ours & \textbf{39.3} \\
			
			\hline
		\end{tabular}
	\end{center}
	\caption{Comparison with the state-of-the-art mesh-based 3D human estimation methods on Human3.6M test set. The numbers are joint errors in mm with Procrustes alignment under P2, and lower is better. 
		Our approach achieves the state-of-the-art performance. }
	\label{h36m}
	\vspace{-5pt}
\end{table}

\begin{table}
	\begin{center}
		\begin{tabular}{c|c}
			\hline
			Methods & Surface Error\\
			\hline
			SMPLify++~\cite{lassner2017unite} & 75.3 \\
			Tung\etal~\cite{NIPS2017_7108} & 74.5 \\
			BodyNet\cite{varol2018bodynet} & 73.6 \\
			\hline
			Ours & \textbf{56.5} \\
			\hline
		\end{tabular}
	\end{center}
	\caption{Comparison with the state-of-the-art methods on SURREAL dataset. The numbers are the mean vertex errors in mm, and lower is better. Our methods outperform baselines with a large margin.}
	\label{surreal}
	\vspace{-5pt}
\end{table}

\begin{table}
	\begin{center}
		\begin{tabular}{c|c|c|c|c}
			\hline
			& \multicolumn{2}{|c|}{FB Seg.} & \multicolumn{2}{|c}{Part Seg}\\
			& acc. & f1 & acc. & f1\\
			\hline
			SMPLify ${oracle}$~\cite{bogo2016keep} & \textbf{92.17} & \textbf{0.88} & 88.82 & 0.67 \\
			SMPLify~\cite{bogo2016keep}  & 91.89 & \textbf{0.88} & 87.71 & 0.67 \\
			SMPLify on \cite{pavlakos2018learning} & 92.17 & \textbf{0.88} & 88.24 & 0.64 \\
			\hline
			HMR~\cite{kanazawa2018end} & 91.67 & 0.87 & 87.12 & 0.60 \\
			CMR~\cite{kolotouros2019convolutional} & 91.46 & 0.87 & 88.69 & 0.66 \\	SPIN~\cite{kolotouros2019learning} & 91.83 & 0.87 & 89.41 & 0.68 \\
			\hline
			Ours & 92.10 & \textbf{0.88} & \textbf{89.45} & \textbf{0.69} \\
			\hline
		\end{tabular}
	\end{center}
	\caption{Comparison with the state-of-the-art methods on LSP test set. The numbers are accuracy and f1 scores, and higher is better.
		SMPLify~\cite{bogo2016keep} is optimization based, while HMR~\cite{kanazawa2018end}, CMR~\cite{kolotouros2019convolutional}, SPIN~\cite{kolotouros2019learning} and our method are regression based. Our framework achieves the state-of-the-art result among regression based methods and is competitive with optimization based methods.}
	\label{lsp}
	\vspace{-10pt}
\end{table}

\subsection{Comparison with the state-of-the-art}
In this section, we present comparison of our method with other state-of-the-art mesh-based methods.

Table \ref{h36m} shows the results on Human3.6M test set. 
We train our model following the setting of CMR~\cite{kolotouros2019convolutional} and  utilize Human3.6M and UP-3D as the training set. 
Our method achieves the state-of-the-art performance among the mesh-based methods. 
It's worth notice that SPIN~\cite{kolotouros2019learning} and our method focus on different aspect and are compatible. SPIN~\cite{tekin2017learning} focus on the training using data with scarce 3D ground truth and the network is trained with extra data from 2D human pose benchmarks. While we focus on the dense correspondence between mesh and image, and do not include data from 2D human pose benchmarks.
%

Similarly, we show the results on SURREAL dataset in Table \ref{surreal}. Our model is trained only with training data of SURREAL dataset and outperforms the previous methods by a large margin. The human shape in SURREAL dataset is of great variety, and this verifies the human shape reconstruction capability of our method.

We also investigate human shape estimation accuracy by evaluating the foreground-background and part-segmentation performance on the LSP test set. 
During the evaluation, we use the projection of the 3D mesh as segmentation result. The predicted IUV image is not used in evaluation for fair comparison.
The results are shown in Table~\ref{lsp}. Our regression based method outperforms the state-of-the-art regression based methods and is competitive with the optimization based methods,
which tend to outperform the regression based methods on this metric but are with much lower inference speed.


\subsection{Ablative studies}

\begin{table}
	\begin{center}
		\begin{tabular}{c|c|c|c|c|c|c|c}
			\hline
			UV & \multirow{2}{*}{$\mathcal{F}_{G}$} & \multirow{2}{*}{$\mathcal{F}_{L}$} & raw & \multicolumn{2}{|c|}{MPJPE} & \multicolumn{2}{|c}{MPJPE-PA} \\
			map & ~ & ~ & pixel & P1 & P2 & P1 & P2 \\
			\hline
			\multirow{4}{*}{SMPL} & \checkmark & & & 72.1 & 68.9 & 51.9 & 49.1 \\
			~ & & \checkmark &  & 71.9 & 69.6 & 47.4 & 44.8 \\
			~ & \checkmark & \checkmark & & 65.0 & 61.7 & 45.1 & 42.6 \\
			~ & \checkmark & & \checkmark & 65.0 & 63.2 & 46.5 & 44.7 \\
			\hline
			\multirow{4}{*}{Ours} & \checkmark & & & 69.5 & 67.7 & 49.4 & 47.1 \\
			~ & & \checkmark &  & 69.8 & 68.4 & 44.6 & 42.3 \\
			~ & \checkmark & \checkmark & & \textbf{62.7} & \textbf{60.6} & \textbf{42.2}  & \textbf{39.3} \\
			~ & \checkmark & & \checkmark & 63.2 & 61.0 & 45.5 & 42.6 \\
			\hline
		\end{tabular}
	\end{center}
	\caption{Comparison on Human3.6M test set with different UV map and input of location net. The numbers are 3D joint errors in mm. $\mathcal{F}_{G}$ and $\mathcal{F}_{L}$ refer to global feature vector and local feature map, respectively. 
	With both UV maps, the framework use local feature outperforms the baseline using global feature with a large margin. Combining global feature and local feature further improves the performance. However, transferring raw image pixels brings a gain much smaller.
	With the same input, the frameworks using our UV map outperform these using SMPL default UV map.
	\label{tab:ablation}
	}
	\vspace{-20pt}
\end{table}

\begin{figure}[t]
	\centering
	\includegraphics[width=0.9\linewidth]{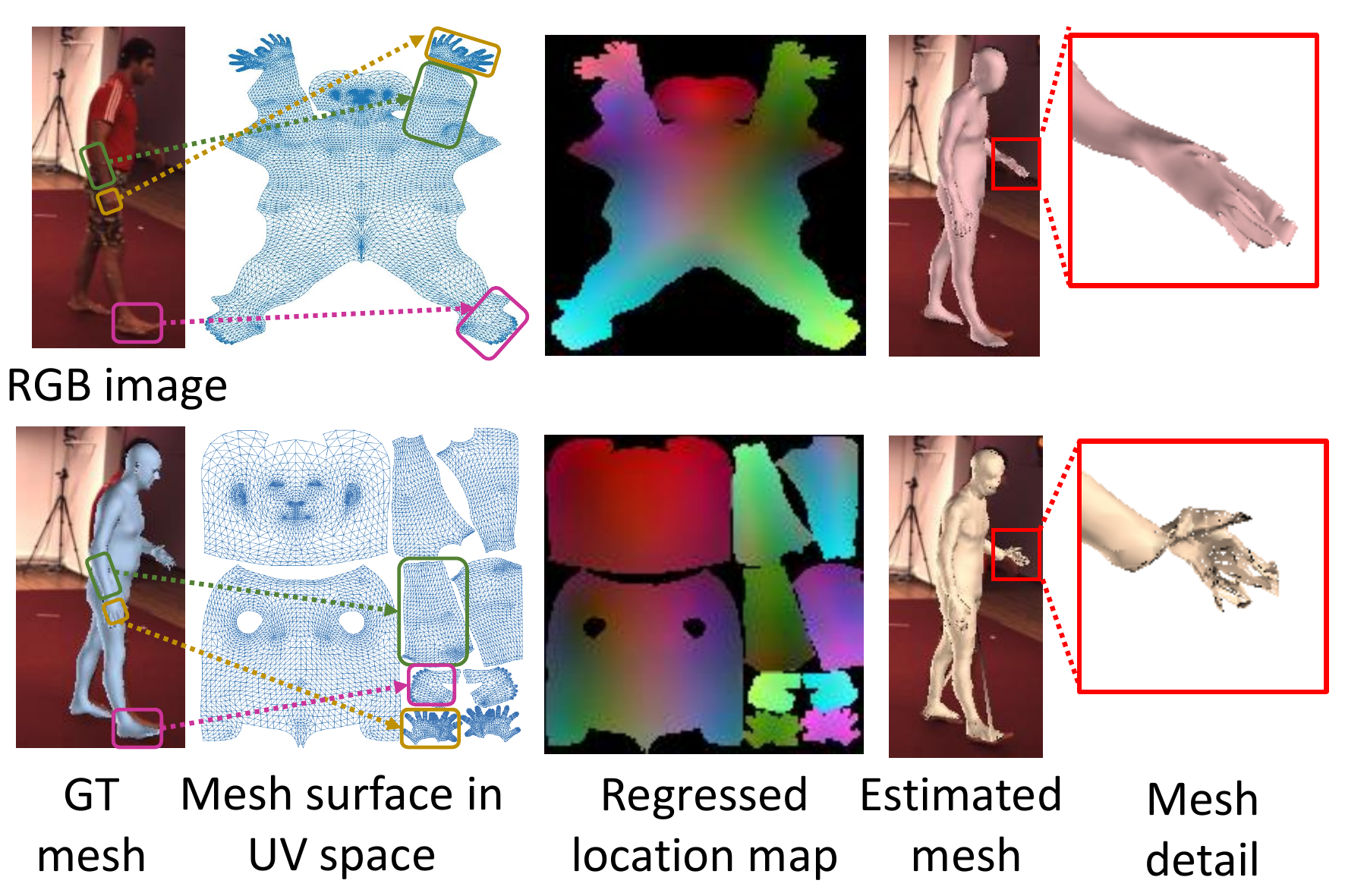}
	\caption{An example of mesh reconstructed using our new UV map (top) and SMPL default UV map (bottom). SMPL default UV map may cause discontinuity between different parts as well as erroneous estimation of some vertices near part edges. While our new UV map mitigates these problems.}
	\label{uv_space}
	\vspace{-15pt}
\end{figure}

\begin{figure*}[t]
	\centering
	\includegraphics[width=0.15\linewidth]{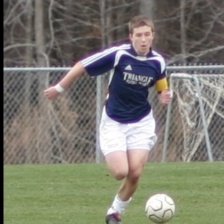}
	\includegraphics[width=0.15\linewidth]{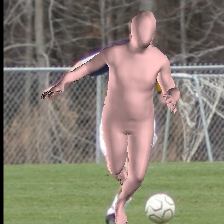}
	\includegraphics[width=0.15\linewidth]{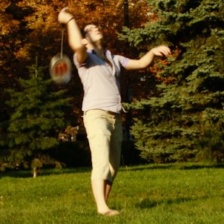}
	\includegraphics[width=0.15\linewidth]{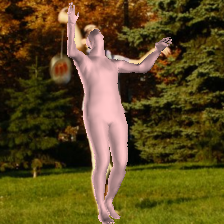}
	\includegraphics[width=0.15\linewidth]{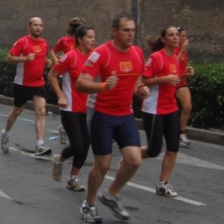}
	\includegraphics[width=0.15\linewidth]{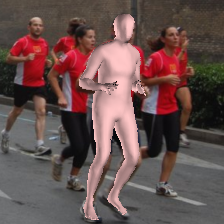}
	
	\includegraphics[width=0.15\linewidth]{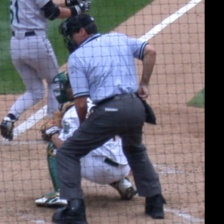}
	\includegraphics[width=0.15\linewidth]{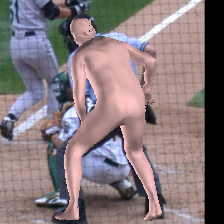}
	\includegraphics[width=0.15\linewidth]{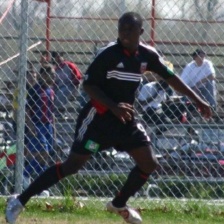}
	\includegraphics[width=0.15\linewidth]{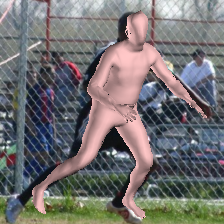}
	\includegraphics[width=0.15\linewidth]{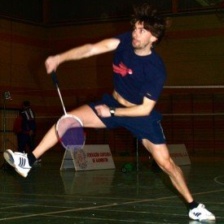}
	\includegraphics[width=0.15\linewidth]{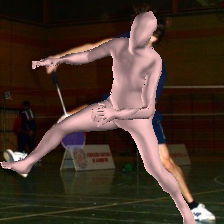}
	
	\includegraphics[width=0.15\linewidth]{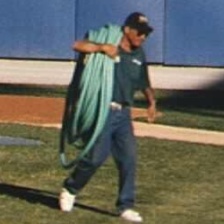}
	\includegraphics[width=0.15\linewidth]{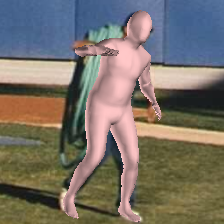}
	\includegraphics[width=0.15\linewidth]{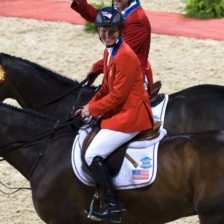}
	\includegraphics[width=0.15\linewidth]{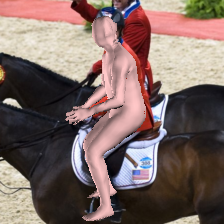}
	\includegraphics[width=0.15\linewidth]{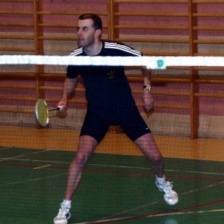}
	\includegraphics[width=0.15\linewidth]{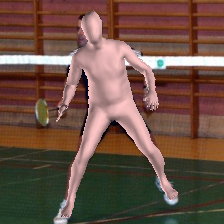}
	
	\includegraphics[width=0.15\linewidth]{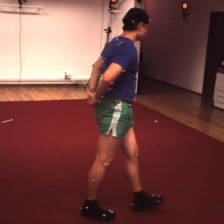}
	\includegraphics[width=0.15\linewidth]{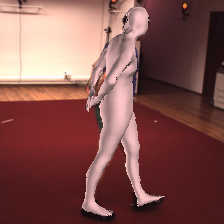}
	\includegraphics[width=0.15\linewidth]{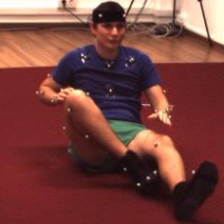}
	\includegraphics[width=0.15\linewidth]{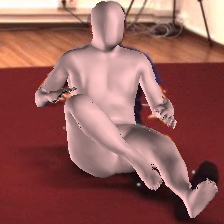}
	\includegraphics[width=0.15\linewidth]{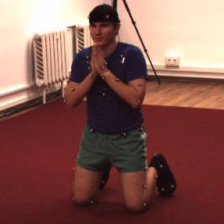}
	\includegraphics[width=0.15\linewidth]{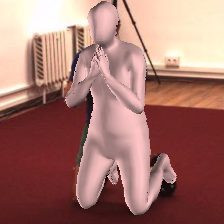}
	
	\includegraphics[width=0.15\linewidth]{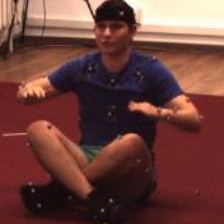}
	\includegraphics[width=0.15\linewidth]{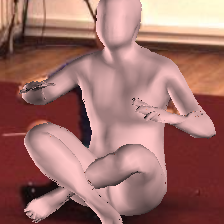}
	\includegraphics[width=0.15\linewidth]{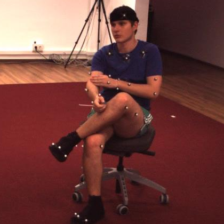}
	\includegraphics[width=0.15\linewidth]{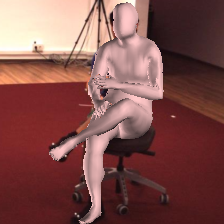}
	\includegraphics[width=0.15\linewidth]{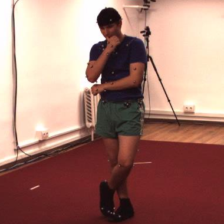}
	\includegraphics[width=0.15\linewidth]{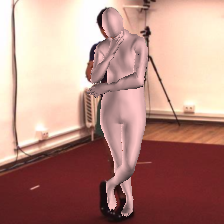}
	
	\caption{Qualitative results of our approach. Rows 1-3: LSP~\cite{Johnson10}. Rows 4-5: Human3.6M~\cite{ionescu2013human3}.}
	\label{qualitative}
	\vspace{-12pt}
\end{figure*}

\begin{figure}[t]
	\centering
	\subfigure[Image]{
		\begin{minipage}[b]{0.20\linewidth}
			\includegraphics[width=1\linewidth]{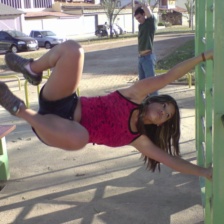}
			\includegraphics[width=1\linewidth]{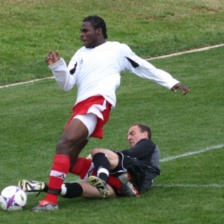}
	\end{minipage}}
	\subfigure[Result]{
		\begin{minipage}[b]{0.20\linewidth}
			\includegraphics[width=1\linewidth]{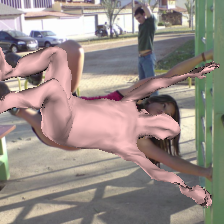}
			\includegraphics[width=1\linewidth]{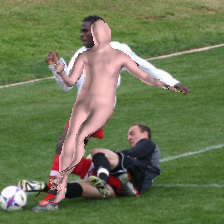}
	\end{minipage}}
	\subfigure[Image]{
		\begin{minipage}[b]{0.20\linewidth}
			\includegraphics[width=1\linewidth]{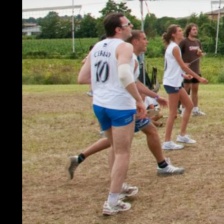}
			\includegraphics[width=1\linewidth]{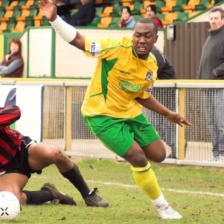}
	\end{minipage}}
	\subfigure[Result]{
		\begin{minipage}[b]{0.20\linewidth}
			\includegraphics[width=1\linewidth]{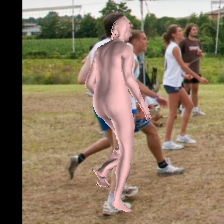}
			\includegraphics[width=1\linewidth]{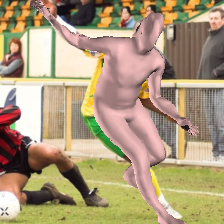}
	\end{minipage}}
	\caption{Examples of erroneous reconstruction of our methods. Typical failures can be attributed to challenging poses, viewpoints rare seen in training set, severe self-osculation, as well as confusion caused by interaction among multiple people.}
	\label{erroneous}
	\vspace{-15pt}
\end{figure}

In this section, we provide the ablation studies of the proposed method.
We train all networks with training data from Human3.6M and UP-3D dataset, and evaluate the models on Human3.6M test set. 

\textbf{Dense correspondence}: 
We first investigate the effectiveness of the dense correspondence between 3D mesh and image features.
We train networks that only use global feature or transferred local feature as the input of LNet.
The comparison is shown in Table \ref{tab:ablation}.
With both UV maps, the framework utilizing transferred local feature outperforms the baseline using global feature with a large margin, which proves the effectiveness of the established dense correspondence. 
Combining global feature with local feature further improves the performance.

We also train frameworks that transfer raw image pixels rather than image features and observe much less improvement than transferring local features.
We attribute this phenomenon to the lack of human pose information in transferred raw pixels.
For images with the same person in different poses, the pixels of a certain body part will be transferred to the same position in the UV space, which generates similar inputs for the LNet. 
So the LNet can only use transferred pixels to refine the estimation of human shape, and predict human pose only based on global feature.

On the contrary, the CNet is able to embed human pose information into image features. Then the LNet can resort to transferred features to refine both human shape and pose estimation.

\textbf{UV map}:
For the second ablative study, we investigate the influence of different UV maps.
We compare the performance of frameworks using SMPL default UV map~\cite{loper2015smpl}, and our continuous UV map. 

As shown in Table~\ref{tab:ablation}, 
with the same input of LNet, the frameworks using our continuous UV map outperforms these frameworks using SMPL default UV map with a large margin.
We attribute the gain to the continuity of the new UV map. 
%
As shown in Figure~\ref{uv_space}, some neighboring parts on mesh surface are distant on SMPL default UV map, such as arms and hands. This may lead to discontinuity of these parts on the final 3D mesh. 
Additionally, some faraway surface parts are very close on the UV plane, such as hands and foots, which might cause erroneous estimation of vertices on edges of these parts. These phenomenons are both shown in Figure~\ref{uv_space}.
On the contrary, our UV map preserves more neighboring relations of the original mesh surface, so these problems are mitigated.

\subsection{Qualitative result}
Some qualitative results are presented in Figure~\ref{qualitative}, and Figure~\ref{erroneous} includes some failure cases. Typical failure cases can be attributed to challenging poses, viewpoints rare seen in training set, severe self-osculation, as well as confusion caused by interaction among multiple people.

\section{Conclusion}
This work aims to solve the problem of lacking dense correspondence between the image feature and output 3D mesh in mesh-based monocular 3D human body estimation. 
The correspondence is explicitly established by IUV image estimation and image feature transferring. 
Instead of reconstructing human mesh from global feature, our framework is able to make use of extra dense local features transferred to the UV space. 
To facilitate the learning of frame work, we propose a new UV map that maintains more neighboring relations of the original mesh surface.
Our framework achieves state-of-the-art performance among 3D mesh-based methods on several public benchmarks.
Future work can focus on extending the framework to the reconstruction of surface details beyond existing human models, such as cloth wrinkles and hair styles.

\section*{Acknowledgement}
We thank reviewers for helpful discussions and comments. 
Wanli Ouyang is supported by the Australian Research Council Grant  DP200103223.
This work is supported in part by the General Research Fund through the Research Grants Council of Hong Kong under Grants (No. 14203118, 14208619), in part by Research Impact Fund Grant No. R5001-18

{\small
	\bibliographystyle{ieee}
	\bibliography{egbib}
}

\clearpage


\section*{Supplementary material}

# 3D Human Mesh Regression with Dense Correspondence
# **Supplementary Material**

Wang Zeng[1], Wanli Ouyang[2], Ping Luo[3], Wentao Liu[4], and Xiaogang Wang[1,4]

[1]The Chinese University of Hong Kong  [2]The University of Sydney  [3]The University of Hong Kong  [4]SenseTime Research
{zengwang@link, xgwang@ee}.cuhk.edu.hk, wanli.ouyang@sydney.edu.au, pluo@cs.hku.hk,
liuwentao@sensetime.com

This supplementary material provides details not included in the main manuscript because of the space constrain. In Section 1, we present the training data used by different methods mentioned in the paper. In Section 3, we provide details about the weight map used in the computation of $\mathcal{L}_{map}$ in the main manuscript. In Section 2, we show some qualitative results on the SURREAL [18] test set and present qualitative comparison between our method and other state-of-the-art mesh-based methods.

## 1. Training data

As mentioned in the Section 4.2 of the main manuscript, the mesh-based methods we mentioned utilize different training data and the results are not directly comparable. In this section, we provides more details about the training data of these methods. We first introduce the related datasets bellow.

**LSP-extended**: LSP-extended [6] is a 2D human pose benchmarks containing 10,000 images with challenging human poses. For every image, 14 visible joint locations are annotated.

**MPII**: MPII [1] is a large scale 2D human pose dataset composed of over 25K images with annotated 2D joint locations. The MPII dataset contains over 40K people and covers 410 human activities.

**MS COCO**: For MS COCO [12], only the part of keypoints detection task is used, which contains over 150,00 people and 1.7 million annotated 2D keypoints.

**MPI-INF-3DHP**: MPI-INF-3DHP [14] is a recent 3D human pose estimation dataset captured by using a multi-view setup and synthetic data augmentation. For each image, ground-truth 3D keypoints locations are provided.

**MOCA**: MOCA [20] is a recent synthetic dataset including 2 million synthetic images with corresponding ground-truth 3D human body shapes and poses.

In Table 1, we present the training data used by each method when evaluated on the Human3.6M [4] test set. Pavlakos *et al.* [16] uses no training data from Human3.6M and trains 3D prior net using data from CMU MoCap [3], while NBF [15] only uses training data from Human3.6M. HMR [7], SPIN [9] and DenseRac [20] all utilize extra training data from 2D human pose benchmarks. SPIN and DenseRac additionally includes training data from the MPI-INF-3DHP dataset [14]. In addition, DenseRac makes use of synthetic data from MOCA [20]. Our method follows the setting of CMR [10], and uses training data from Human3.6M and UP-3D [11] without extra data from 2D human pose benchmarks. Our framework outperforms CMR with a large margin on the Human3.6M test set (the MPJPE of CMR and our method are 50.1 mm and 39.3 mm respectively).

Although SPIN and our method have similar performance on Human3.6 test set, the contributions are totally different. The impressive performance of SPIN can be attributed to its effective utilization of training data from 2D human pose benchmarks. However, our method focuses on the dense correspondence between 3D mesh and image, as well as the utilization of local image features. Therefore, SPIN and our method are complementary.

## 2. Qualitative results

In this section, we present some qualitative results of our method. Figure 1 shows some qualitative results of our method on the SURREAL [18] test set. Our method is able to reconstruct 3D human bodies with various shapes and poses.

Figure 2 shows some qualitative results of our method and other state-of-the-art methods on the test set of Human3.6M [4]. The state-of-the-art model-free method (*i.e.* CMR [10]) and model-based method (*i.e.* SPIN [9]) all estimate the full human body based on the global image feature extracted by CNN and may fail to reconstruct details which

| Datasets | Pavlakos *etc.* [16] | NBF[15] | HMR [7] | SPIN [9] | DenseRac [20] | CMR [10] | Ours |
|---|---|---|---|---|---|---|---|
| Human3.6M [4] | | ✓ | ✓ | ✓ | ✓ | ✓ | ✓ |
| LSP [5] | ✓ | | ✓ | ✓ | ✓ | | |
| LSP-extended [6] | ✓ | | ✓ | ✓ | | | |
| MPII [1] | ✓ | | ✓ | ✓ | ✓ | | |
| MS COCO [12] | | | ✓ | ✓ | ✓ | | |
| MPI-INF-3DHP [14] | | | | ✓ | ✓ | | |
| MOCA [20] | | | | | ✓ | | |
| UP-3D [11] | | | | | | ✓ | ✓ |

Table 1. The training data used by different methods when evaluated on the Human3.6M [4] test set. Our approach uses the same training data with CMR [10].

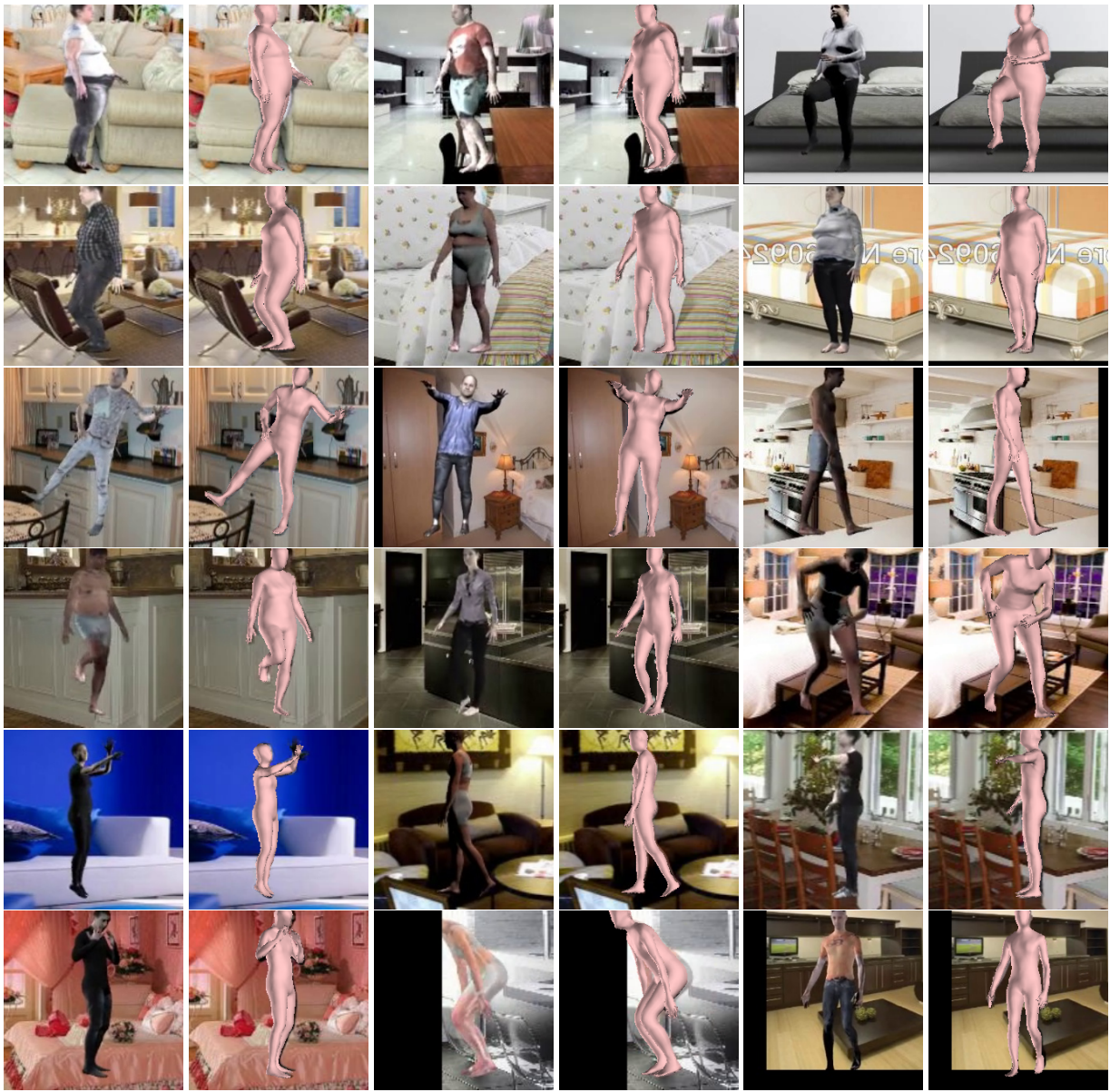

Figure 1. Qualitative results of our approach on the SURREAL [18] test set.

12

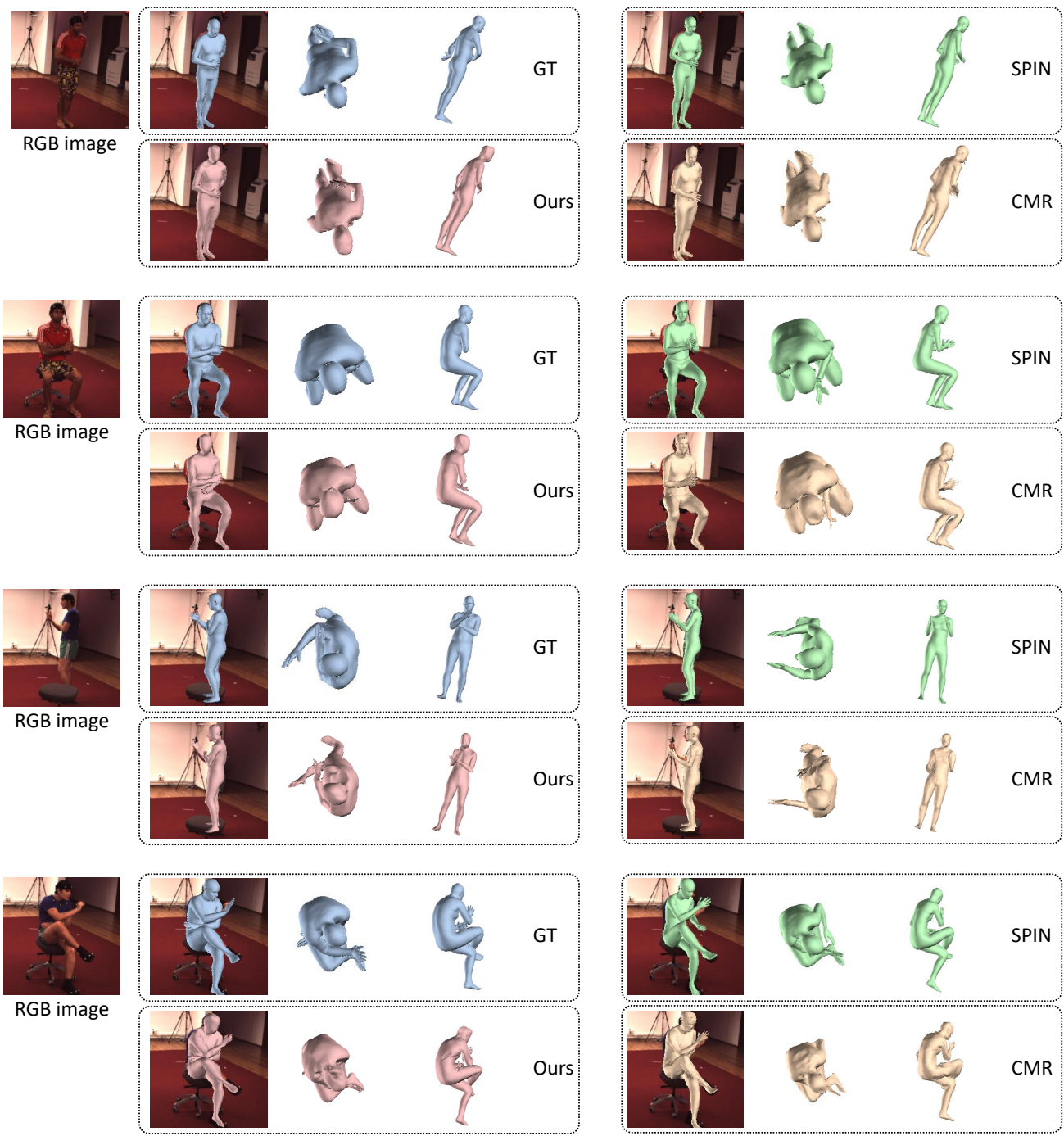

Figure 2. Comparison between our method and other state-of-the-art 3D mesh-based methods. CMR [10] and SPIN [9] may fail to reconstruct details which are not distinct on the image, while our method is able to reconstruct these details well.

are not distinct on the image. However, our method can utilize local image features with the explicitly established correspondence between mesh and image, and is able to reconstruct these details better.

## 3. Weight map

This section introduces the weight map for the loss term between regressed location map and ground-truth location map (*i.e.* $\mathcal{L}_{map}$). We assign larger weights to the parts away

13

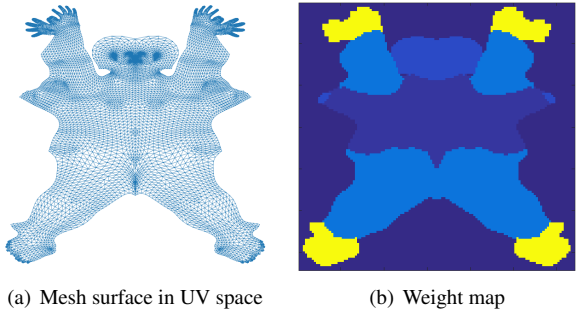

(a) Mesh surface in UV space  (b) Weight map

Figure 3. Illustration of the weight map used for $\mathcal{L}_{map}$. Surface parts away from torso are assigned with larger weights.

from the torso, whose locations are with larger variance and more difficult for the network to estimate.

Specifically, we generate the weight map using the mesh part segmentation provided by SMPL model [13]. Different weights are assigned to different surface parts to get the weight map. Denote the weights for torso, neck&head, arms&legs, hands&foots respectively as $\lambda_t$, $\lambda_{n\&h}$, $\lambda_{a\&l}$, $\lambda_{h\&f}$. We set $\lambda_t : \lambda_{n\&h} : \lambda_{a\&l} : \lambda_{hs\&f}$ as $1 : 2 : 5 : 25$. Figure 3 shows the normalized weight map.

## 4. Evaluation on 3DPW dataset

3DPW [19] dataset is a recent outdoor 3D human body estimation benchmark. It provides 3D human pose and shape ground truth captured with IMU sensors. In our work, we only use its test set for evaluation.

| Method | MPJPE-PA |
|--------|----------|
| HMR    | 81.3     |
| CMR    | 70.2     |
| [8]    | 72.6     |
| [2]    | 72.2     |
| [17]   | 69.9     |
| Ours-A | **68.5** |
| Ours-B | 61.7     |
| SPIN   | **59.2** |

Table 2. Comparison with the state-of-the-art methods on 3DPW. SPIN and Ours-B utilize fitted SMPL parameters from SPIN for training, while other methods do not. Without using fitted SMPL parameters, our framework outperforms the methods using only global features. Utilizing fitted SMPL parameters further improves the performance to be competitive with the state-of-the-art method.

In order to investigate the generalization capability of our method, we evaluate our method on 3DPW test set. We use extra data from COCO [12], LSP [5] and MPII [1] as weak supervision to scale up our model (Ours-A) for fair comparison with prior works. We also train our model (Ours-B) with part of the fitted SMPL parameters from SPIN.

The results are presented in Table 2. Without using fitted SMPL parameters, our model outperforms the methods using only global features. Utilizing fitted SMPL parameters further improves the performance to be competitive with the state-of-the-art. It is worth notice that we did not include the training data of LSP-extended and MPI-INF-3DHP as SPIN. Combining our method with the in-the-loop optimization process in SPIN may bring further performance improvement.



\end{document}